\documentclass[conference]{IEEEtran}
\IEEEoverridecommandlockouts
\usepackage{cite}
\usepackage{amsmath,amssymb,amsfonts}
\usepackage{algorithmic}
\usepackage{graphicx}
\usepackage{subcaption}
\usepackage{textcomp}
\usepackage{xcolor}
\usepackage{tikz}
\usetikzlibrary{positioning}
\def\BibTeX{{\rm B\kern-.05em{\sc i\kern-.025em b}\kern-.08em
    T\kern-.1667em\lower.7ex\hbox{E}\kern-.125emX}}
\begin{document}

\title{Predictive Mapping of Spectral Signatures from RGB Imagery for Off-Road Terrain Analysis}

\author{Sarvesh Prajapati$^{1}$, Ananya Trivedi$^{1}$, Bruce Maxwell$^{2}$, Ta\c{s}k{\i}n Pad{\i}r$^{1}$
\thanks{$^{1}$Institute for Experiential Robotics, Northeastern University, Boston, Massachusetts, USA. { \tt\small \{prajapati.s, trivedi.ana, t.padir\}@northeastern.edu}}
\thanks{$^{2}$Khoury College of Computer Sciences, Northeastern University, Boston, Massachusetts, USA. { \tt\small b.maxwell@northeastern.edu}}

}
\maketitle
\begin{abstract}

Accurate identification of complex terrain characteristics, such as soil composition and coefficient of friction, is essential for model-based planning and control of mobile robots in off-road environments. Spectral signatures leverage distinct patterns of light absorption and reflection to identify various materials, enabling precise characterization of their inherent properties. Recent research in robotics has explored the adoption of spectroscopy to enhance perception and interaction with environments. However, the significant cost and elaborate setup required for mounting these sensors present formidable barriers to widespread adoption. In this study, we introduce RS-Net (RGB to Spectral Network), a deep neural network architecture designed to map RGB images to corresponding spectral signatures. We illustrate how RS-Net can be synergistically combined with Co-Learning techniques for terrain property estimation. Initial results demonstrate the effectiveness of this approach in characterizing spectral signatures across an extensive off-road real-world dataset. These findings highlight the feasibility of terrain property estimation using only RGB cameras.
\end{abstract}

\section{Introduction}
Understanding terrain properties is vital in off-road robotics, as they directly shape the robot's performance and abilities \cite{ram_vasudevan_terrain_estimation}. These properties encompass factors such as soil composition, friction coefficients, obstacle density, among other variables, all significantly influencing navigation, stability, and obstacle interaction. By comprehensively understanding and adapting to these terrain properties, robots can ensure safe and efficient traversal through diverse and challenging outdoor environments\cite{hanson_hyper_data,hanson_thesis}.

Recent trends in off-road autonomy involve estimating terrain properties and incorporating this information into a motion planning algorithm \cite{mppi,icra_ananya,ram_vasudevan_terrain_estimation,icra}. In a basic approach, terrain classification is first conducted using an off-the-shelf classifier, followed by using a lookup table to estimate the properties associated with each class \cite{brandao_segmentation_for_friction}. However, this method loses generality when it comes to distinguishing between similar-looking materials or dealing with previously unseen objects. Another trend involves the use of specialized sensors such as haptics \cite{haptics}, lidars \cite{evora} and thermal sensors \cite{thermal}, where physical interaction with materials is utilized to estimate their properties. While these approaches provide valuable insights, they require dedicated sensors and involve physically touching objects to estimate their properties.

\begin{figure}[t]
\centering
\subfloat{\tikz[remember picture]{\node(1AL){\includegraphics[width=0.25\linewidth]{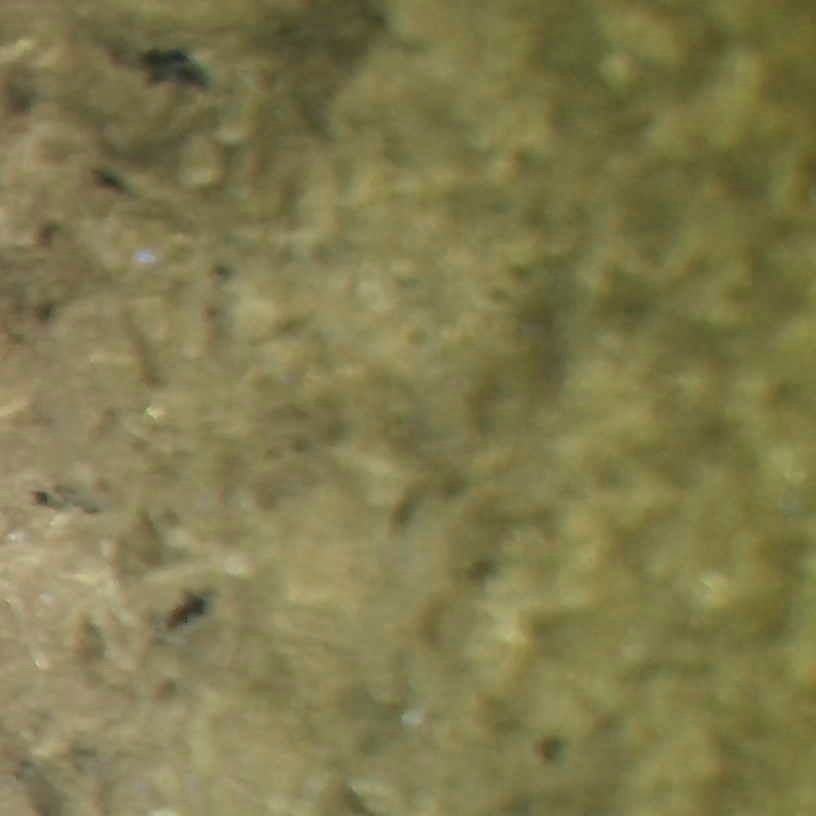}};}\vspace{3.5em}}%
\hspace*{0.5cm}%
\subfloat{\tikz[remember picture]{\node(1AR){\includegraphics[width=0.7\linewidth]{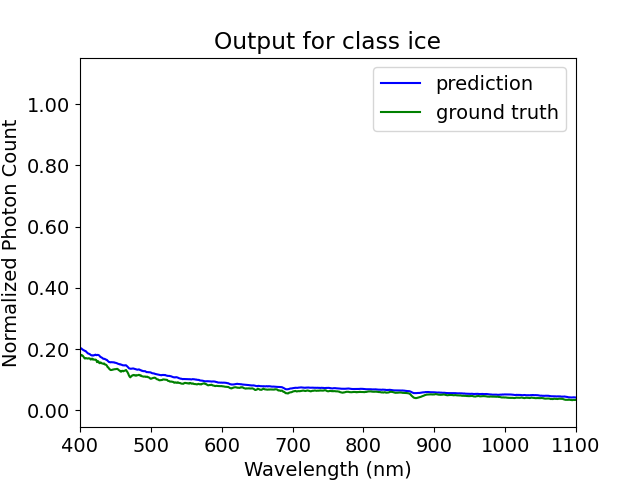}};}}
\label{fig:intro}
\caption{Proposed RS-Net takes in RGB image and accurately predicts spectral value of a material patch.}
\end{figure}
\tikz[overlay,remember picture]{\draw[-latex,thin] (1AL) -- (1AL-|1AR.west)
node[midway,below,text width=1.5cm]{};}

Spectroscopy is a scientific technique for examining how light interacts with materials \cite{what_is_spectroscopy}. It measures the spectrum of light that materials emit, absorb, or scatter. This approach enables learning about the material's composition, structure, and chemical properties. Each type of material has a unique pattern of wavelengths, known as a spectral signature, which is revealed when light interacts with it. By studying these signatures, spectroscopy provides a powerful tool for understanding the properties of different substances \cite{spec_1,spec_2}. Additionally, spectroscopy enables the observation of temporal changes in object properties, with alterations in electromagnetic radiation absorption or emission serving as indicators of ongoing compositional or physical transformations.

\begin{figure*}[t]
    \centering
    \includegraphics[width=0.98\linewidth]{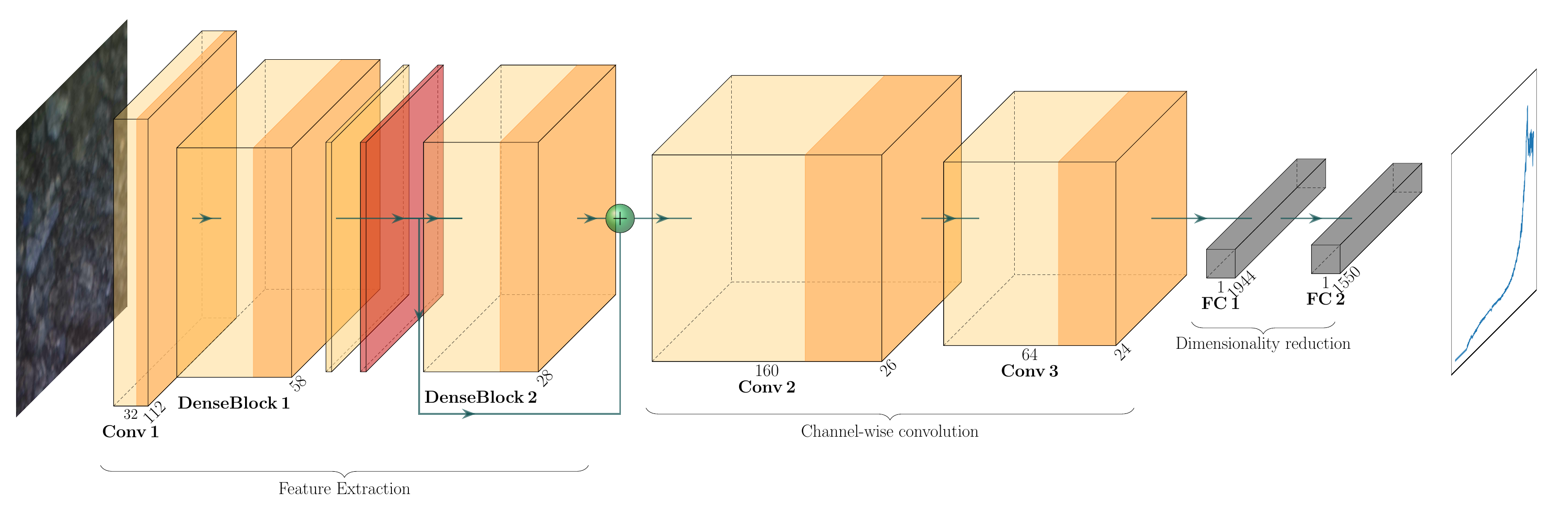}
    \caption{RS-Net: End-to-End model for mapping RGB images to spectral profile. The input RGB image is passed through DenseNet's first and second dense block, output from first transitional layer is concatenated with output of second dense block, the feature maps are concatenated, passed through convolution and fully-connected layers for spectral mapping.}
    \label{fig:efnet}
\end{figure*}

In our work, we propose utilizing spectroscopy to detect terrain properties based on their unique spectral signatures. These sensors, specifically designed for estimating physical properties, offer high accuracy. We demonstrate how a supervised deep neural network architecture can accurately predict spectral signatures from RGB data. Additionally, we illustrate how these spectral signatures can be leveraged to estimate various terrain properties, such as surface friction. This will eliminate the requirement for custom sensor assemblies currently needed in robotics spectroscopy applications \cite{slurp,hanson_thesis}. Our approach aims to establish a framework where offline-collected spectral data and a trained network can be used to estimate any terrain physical property, offering a more generalized solution compared to the custom solutions commonly employed in current terrain property estimation algorithms.

\section{Network Architectures}
In this section, we introduce RS-Net, a deep neural network architecture to convert RGB images into a spectral profile. Furthermore, we put forward a potential methodology for expanding this architecture to facilitate the estimation of terrain properties. 

\subsection{RS-Net Architecture}

Figure~\ref{fig:efnet} illustrates the RS-Net architecture. Figure~\ref{fig:seen} shows a typical spectral profile that indicates how the intensity of light varies across the spectrum, providing information about the wavelengths of light that are present and their relative intensities. RS-Net takes an input RGB image $x$ and predicts a spectral profile $x'_s$. In the following sections, we break down the various components integral to our neural network design.

\subsubsection*{DenseNet}
To extract high-level global features, the input RGB image undergoes multi-stage processing. We utilize a pretrained DenseNet169 \cite{DBLP:journals/corr/HuangLW16a} for this purpose, with the flexibility to substitute other DenseNet models, as the initial two blocks remain consistent across variants. In DenseNet, the image is initially processed through a convolution and pooling layer. We represent this as a composition of operations $G(\cdot)$.
\begin{equation} \label{eq:conv_and_pooling}
    x_1 = G(x)
\end{equation}
The input image is then processed through dense block 1 and 2 of the DenseNet. This sequential passage enables the network to learn fine-grained details in dense block 1 and high-level abstract representations in dense block 2. Features extracted from these layers contribute to capturing the texture of the surface, thereby enhancing our ability to achieve better spectral correspondence and differentiating between materials of similar appearance. The first dense block has 6 layers of convolution while the second has 12 layers of convolutions. After first dense block, we get $x_7$, which is passed through transition layer that performs operation $G(\cdot)$. Finally this output from transitional layer is passed through second dense block, outputting $x_{21}$.
As per DenseNet, the general representation of a dense block for $l$ convolutions is:
\begin{equation}
    x_l = H_l(\left[x_0, x_1, ..., x_{l-1}\right])
\end{equation}
Which for our network cam be written as:
\begin{align} \label{eq:dense_block_1_and_2}
    x_7 &= H_7(\left[x_1, x_2,...,x_6\right]) \\
    x_8 &= G(x_7) \\
    x_{21} &= H_{21}(\left[x_8, x_9,...,x_{20}\right])
\end{align}

\subsubsection*{Fused Representation}
Features from max-pooling of first transition layer and second dense block are concatenated for creating a fused representation of the global features. This operation outputs $x_f$ which is the concatenation of $x_8$ and $x_{21}$ and represented as:
\begin{equation}
    x_f = x_8 \oplus x_{21}
\end{equation}

\subsubsection*{CNN Layers}
The fused representation is then fed through the first CNN layer, reducing input channels from 160 to 64, which in-turn helps in reducing the spatial dimension.

Subsequently, the output from the first CNN layer is processed by a second CNN layer, with 64 input channels and 9 output channels. This step is done for regularization, and reducing the dimensionality such that when the network is passed through fully-connected layers, it is more memory efficient and learns the spectral representation efficiently.

\subsubsection*{Fully Connected Layers}
The output from the second CNN layer is flattened and passed through a fully connected layer with input channels as 1944 and output channels as 1550.

This is followed by another fully connected layer, maintaining input and output channels at 1550. The above can be represented as a series composite function, $S(\cdot)$, and written as:
\begin{equation}
    x'_s = S(x_f)
\end{equation}

This is how our network maps RGB images $x$ to their corresponding spectral profile $x_s^{'}$. We evaluate the performance of this network on a real-world off-road terrain dataset in section~\ref{sec:experiments}.

\subsection{Spectral Profiles to Terrain Property Estimation}
\begin{figure}[hb]
    \centering
    \includegraphics[width=0.98\linewidth]{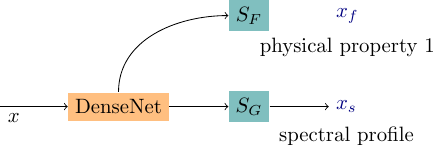}
    \caption{Example of Co-Learning for simultaneous spectral profile and terrain property estimation}
    \label{fig:co-learning}
\end{figure}

We propose an extension of the network to predict various physical properties associated with the terrain based on the given spectral profile $x_s^{'}$. Traditional linking of physical properties with RGB images faces challenges due to the under-constrained nature of the problem \cite{intrinsic_barrow, intrinsic_wise}. Research, such as that by Hanson et al. \cite{hanson2022vast}, highlights the benefits of utilizing different modalities for estimating diverse physical properties crucial for off-road terrain analysis. Instead of treating the task as a straightforward pattern-matching problem, we employ the concept of Collaborative-Learning (Co-Learning) \cite{song2018collaborative}. This approach enables learning representations from multiple modalities, whether they have strong or weak relations, facilitating knowledge transfer between them. By efficiently using RGB to predict spectral data and leveraging the learned parameters to infer a third modality, such as friction, our Co-Learning architecture facilitates the estimation of various physical properties of different off-road materials. Typically, Co-Learning architectures incorporate a split between modalities to predict different representations, facilitating seamless knowledge transfer and enhancing the understanding of underlying physical properties. Figure \ref{fig:co-learning} shows a sample Co-learning architecture for terrain property analysis. The image $x$ is passed through DenseNet, $S_F$ and $S_G$ are composite functions, they can be convolution, max-pooling, batch normalization and even fully-connected layers.

\section{Experiments} \label{sec:experiments}

\subsection{Evaluating RS-Net on trained labels and untrained labels}
\begin{figure}
    \centering
    \includegraphics[width=0.9\linewidth]{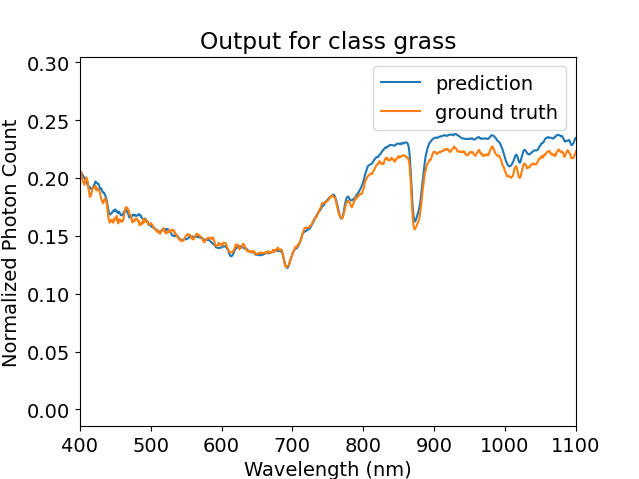}
    \caption{Performance of RS-Net on seen class}
    \label{fig:seen}
\end{figure}

For evaluating our method, we utilize the open sourced dataset from \cite{hanson2022vast} which includes high-resolution macro images of surface texture, spectral reflectance curves, RGB images, and localization data from a 9-DOF IMU, obtained across 11 varied terrains and different lighting conditions. We trained RS-Net on 6 out of the 11 classes in the VAST dataset. These classes include asphalt, brick, grass, ice, sand, and tile. However, the model was not trained on carpet, concrete, gravel, mulch and turf. The model was trained for 50 epochs, on an Adam optimizer \cite{adam} with a learning rate of $1e^{-3}$. Upon evaluation on the testing data, it achieved an MSE loss of $0.0015$ Normalized Photon Count. Figure \ref{fig:seen} displays the output of RS-Net on grass, which constitutes one of the classes the network was trained for. The model demonstrates a remarkable ability to predict the spectral profile with high fidelity.

\begin{figure}
    \centering
    \begin{subfigure}[t]{0.75\linewidth}
        \centering
        \includegraphics[width=0.95\linewidth]{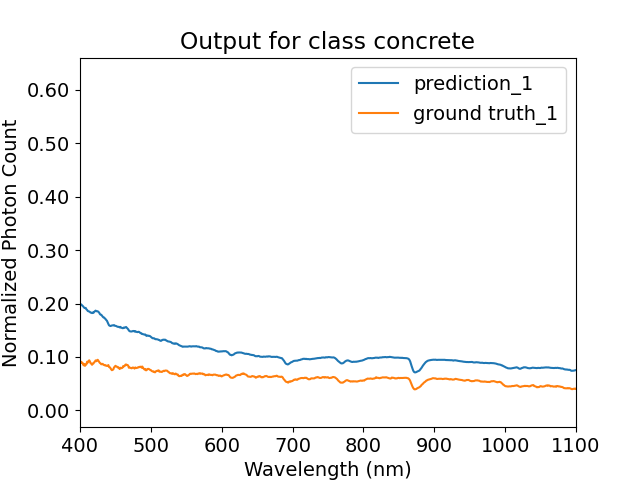}
    \end{subfigure}
    \begin{subfigure}[t]{0.75\linewidth}
        \centering
        \includegraphics[width=0.95\linewidth]{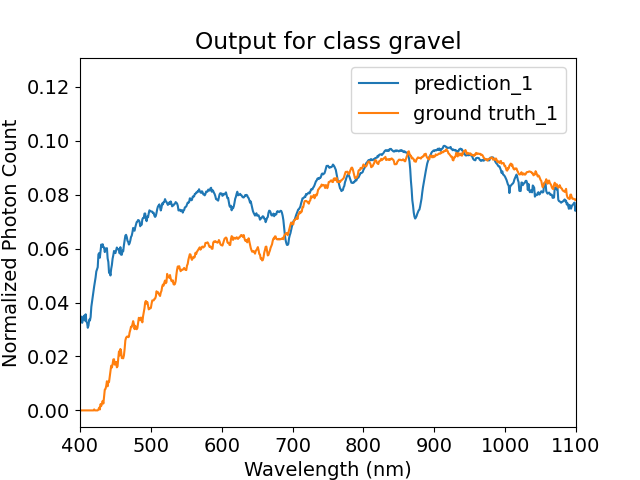}
    \end{subfigure}
    \begin{subfigure}[t]{0.75\linewidth}
        \centering
        \includegraphics[width=0.95\linewidth]{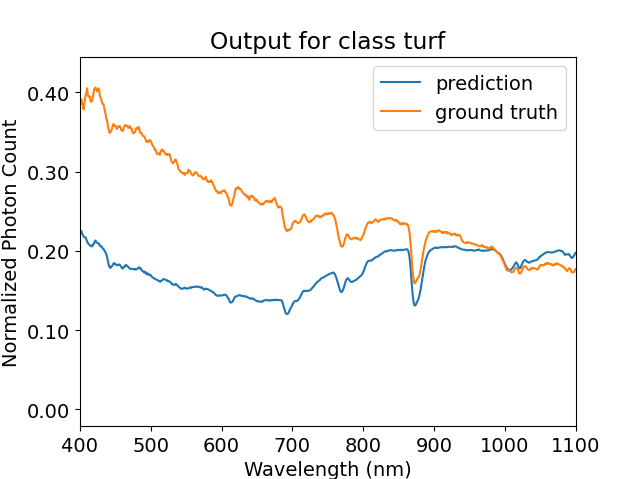}
    \end{subfigure}
    \caption{Predicted spectral profile v/s ground truth data for classes not in training set}
    \label{fig:untrained-class}
    
\end{figure}

Furthermore, we assess RS-Net's performance using data featuring unfamiliar labels to evaluate its capacity for generalization to novel materials. Illustrated in Figure \ref{fig:untrained-class}, the spectral profile of turf, resembling that of grass, produces an output similar to grass, while gravel, akin to sand, and concrete, somewhat akin to asphalt, yield corresponding outputs. This suggests the network's potential to classify diverse objects based on their materials or physical properties, even with limited training. Additional outputs from the network, for both familiar and unfamiliar labels, are elaborated in the Appendix section.

\subsection{Testing material classification}
\begin{figure}
    \centering
    \includegraphics[width=0.90\linewidth]{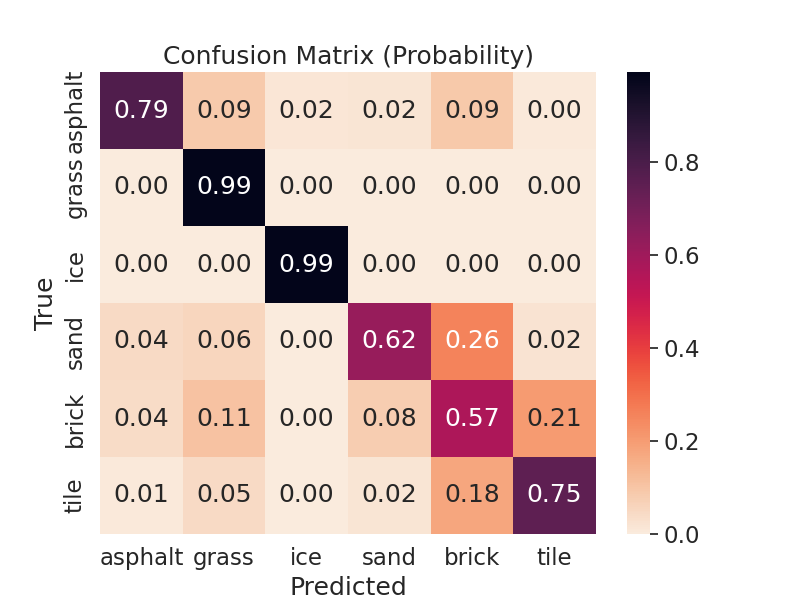}
    \caption{Confusion matrix for terrain classification}
    \label{fig:spectral_asphalt_conf}
\end{figure}

We conduct basic material classification tests using the output of RS-Net. Rather than employing Co-Learning or the concept of perceptual loss, we simply pass the spectral data through a $4$-layer fully connected network and trained it for ten epochs. The confusion matrix on the test data is illustrated in Figure \ref{fig:spectral_asphalt_conf}. The overall F-Score for the testing data was $0.79$ with class accuracy of $0.79, 0.99, 0.99, 0.62, 0.57, 0.75$ for asphalt, grass, ice, sand, brick and tile respectively.

Overall, the network demonstrates an ability to discern patterns even in the case of unseen classes, indicating promising potential for generalization.

\begin{figure}
    \centering
    \includegraphics[width=0.8\linewidth]{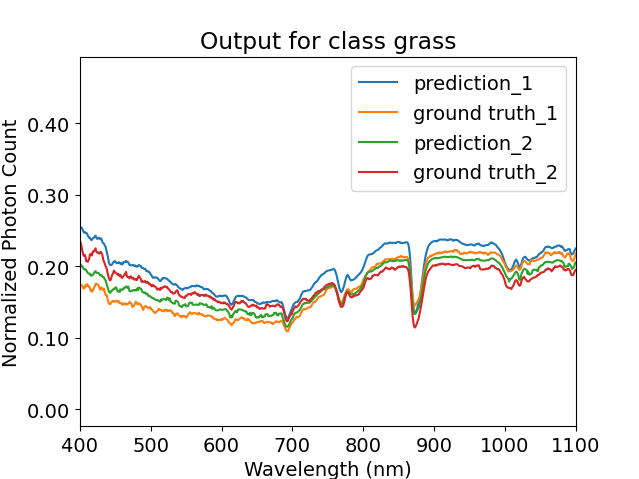}
    \caption{Spectral profile of grass in different environment conditions}
    \label{fig:spectral_asphalt}
\end{figure}

\section{Discussion and Future Work}

In robotics, a key challenge lies in generalizing models for dynamic environments, where existing datasets often fail to capture the variability. For instance, spectral profiles and surface friction can vary based on factors like weather conditions, such as "dry asphalt," "ice on asphalt," and "water on asphalt." Figure \ref{fig:spectral_asphalt} shows how grass's spectral profile shifts in the y-axis with changes in environmental conditions, reflecting corresponding changes in predictions and the network's adaptability to RGB data variations. To enhance model generalization, we intend to enrich the VAST dataset with such nuanced data points.

Our future work involves exploring the relationship between spectral profile shifts and material properties, and developing a Co-Learning network capable of predicting various physical properties like friction and traversability. We also aim to investigate how material deposition affects spectral profiles and, consequently, physical properties. This research direction promises to deepen our understanding of material interactions in dynamic environments and enhance robotic system capabilities.

The existing RS-Net faces deployment challenges as it relies on forwarding material patches to the network, which limits its effectiveness. This dependency on image segmentation or passing patches beneath the robot affects the network's accuracy in long-range forecasting. To overcome this limitation, we suggest incorporating hyperspectral data \cite{hyperspectral} by either transfer learning of spectral data through learning  or adapting the network to directly learn features from hyperspectral data.

\section{Conclusion}

This study underscores the importance of integrating different modalities in material recognition and friction estimation tasks to advance robotic applications. The introduction of RS-Net, a novel architecture mapping RGB images to spectral profiles, presents a promising solution to this challenge. Through experimentation, we have demonstrated the network's ability to generalize to unseen classes, indicating its potential to improve terrain property estimation tasks. Furthermore, we discuss the limitations of current datasets and propose how an augmented data set can enhance the network's ability to learn detailed physical information about a material from just an RGB image.

\section*{Acknowledgement}

The authors would like to thank Dr. Nathaniel Hanson for his help with the VAST \cite{hanson2022vast} dataset.

\bibliographystyle{IEEEtran}
\bibliography{references.bib}

\section*{\textbf{Appendix}}

\subsection{Results}

\begin{figure}[!h]
    \centering
    \includegraphics[width=0.90\linewidth]{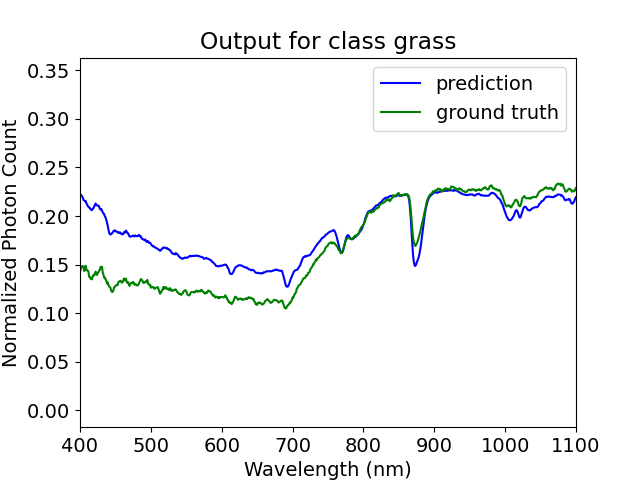}
    \caption{Predicted spectral profile v/s ground truth data for grass}
    \label{fig:grass-append}
\end{figure}

\begin{figure}[!h]
    \centering
    \includegraphics[width=0.90\linewidth]{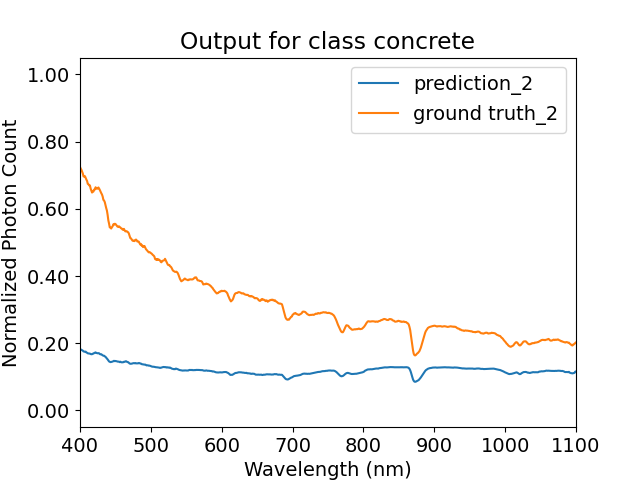}
    \caption{Predicted spectral profile v/s ground truth data for concrete}
    \label{fig:appendix-end}
\end{figure}

Figure \ref{fig:grass-append} and \ref{fig:appendix-end} depict the predicted versus ground truth spectral profiles generated by RS-Net. It is evident that certain materials may exhibit similar profiles at specific wavelengths, suggesting the potential use of this approach to categorize on unknown materials into known categories. A shift in the y-axis can be seen in Figure \ref{fig:grass-append} compared to Figure \ref{fig:seen} suggesting that the data was collected under different environment condition. Figure \ref{fig:appendix-end} demonstrate RS-Net's ability to generalize on out-of-class data. Notably, as the features of asphalt and concrete somewhat overlap, the model produces somewhat noisy and/or shifted profiles for concrete as observed in Figure \ref{fig:appendix-end}.

\end{document}